\documentclass[preprint,12pt]{elsarticle}

\usepackage{graphicx}%
\usepackage{multirow}%
\usepackage{amsmath,amssymb,amsfonts}%
\usepackage{amsthm}%
\usepackage{xcolor}%
\usepackage{textcomp}%
\usepackage{booktabs}%
\usepackage{bm}%
\usepackage{subcaption}%

\usepackage[pagewise, mathlines]{lineno}

\usepackage[colorlinks]{hyperref}
\hypersetup{
  citecolor = {blue}
}

\usepackage{geometry}
\usepackage{caption}
\usepackage{subcaption}
\usepackage{float}
\usepackage{algorithm}
\usepackage{algpseudocode}
\usepackage{tabularx}
\usepackage{url}
\usepackage{multirow}
\usepackage{chngcntr}

\begin{document}

\begin{frontmatter}

\title{Reinforcement Learning-Based Control of CAV Platoon Joining Maneuvers in Mixed Traffic}

\author[Saclay]{Biao Yin\corref{cor1}}
\ead{biao.yin@univ-eiffel.fr}

\author[Dauphine]{Abderrahmane Kasmi}

\author[Marne]{Nadir Farhi}

\cortext[cor1]{Corresponding author.}

\address[Saclay]{Université  Gustave Eiffel,  SATIE, Gif-sur-Yvette, 91190, France.}

\address[Dauphine]{Université Paris Dauphine-PSL, Paris, 75016, France.}

\address[Marne]{Cosys-Grettia, Univ Gustave Eiffel, F-77454 Marne-la-Vallee, France.}

\begin{abstract}
Connected and automated vehicle (CAV) platooning offers a promising approach to improving road safety and traffic capacity. However, platoon control in real-world traffic is challenging due to uncertainty and heterogeneous driving behaviors. Reinforcement learning (RL) has strong potential for addressing such control problems, but its practical deployment raises challenges related to safety and learning efficiency. This paper proposes a generic modeling and simulation framework for investigating CAV platoon joining maneuvers and comparing deep reinforcement learning (DRL)-based control algorithms. The problem is particularly challenging in mixed-traffic environments, where CAVs coexist with human-driven vehicles exhibiting heterogeneous longitudinal and lateral behaviors. The objective is to achieve safe and efficient joining maneuvers by either incorporating penalties for risky behaviors into the learning process or using an external safety controller to constrain the learned policy. An agent-based modeling framework coupled with the Simulation of Urban MObility (SUMO) simulator is used to evaluate Deep Q-Network (DQN), Double Deep Q-Network (DDQN), and Proximal Policy Optimization (PPO). Results show that PPO outperforms DQN and DDQN, achieving a joining success rate of approximately 98 \% and a collision rate below 1 \%, largely due to risk-related penalties incorporated into the reward function. However, this improved performance requires more decision steps to complete the maneuver, revealing a trade-off between safety, joining effectiveness, and decision efficiency. An external safety controller effectively prevents collisions, although its interventions may reduce joining efficiency. The results highlight the importance of jointly considering safety and efficiency when designing RL-based controllers for CAV platoon joining in mixed traffic. The source code is available at: \\
\href{https://github.com/biaoyin/platoon-drl/tree/platoon_join}{https://github.com/biaoyin/platoon-drl/tree/platoon\_join}
\end{abstract}


\end{frontmatter}

\section{Introduction}
\label{intro}
CAVs enable vehicle-to-vehicle and vehicle-to-infrastructure communications to achieve cooperative control and mitigate undesirable human driving behavior (e.g., aggression and stop-and-go instability) \cite{Ahmed2022}\cite{Jiang2021}. The benefits of CAVs extend to traffic flow, yielding improvements in safety and efficiency, as well as contributing to environmental sustainability. Abundant studies have focused on longitudinal speed coordination control of CAVs within a platoon, in which a group of CAVs moves at a consensual speed while maintaining a small, nearly constant distance between adjacent vehicles, such as cooperative adaptive cruise control (CACC). It has been demonstrated that CAV platoons have enormous potential to enhance road capacity \cite{Sala2021}, traffic stability \cite{Yang2025}, and energy efficiency \cite{li2024}. 

Platooning optimization typically involves several key operational phases in the lifecycle of a platoon. These phases structure how vehicles form, operate within, and leave a platoon. Each phase requires sophisticated mechanisms to ensure both safety and efficiency. Although car-following dynamics (such as CACC) have been extensively studied, the initial platoon joining phase is fundamentally important, enabling a CAV to merge into a platoon at a designated position. This requires precise speed control and well-timed lane-changing maneuvers. A key challenge lies in managing longitudinal and lateral control that avoids traffic disturbances, such as congestion or collisions. This challenge is further intensified in mixed traffic, where CAVs coexist with human-driven vehicles in the near future.

In previous work, conventional approaches for platoon joining have been mainly explored based on rules, optimization models, or conceptualized techniques such as virtual platoons \cite{ding2019},\cite{liu2018},\cite{huang2019}. Rule-based approaches enable coordinated merging behavior with low computational complexity, but their performance may be limited in highly dynamic environments \cite{ding2019}. Optimization-based approaches, such as model predictive control \cite{liu2018}, can improve flexibility and performance by formulating the problem as a constrained optimization problem, while requiring higher computational effort and more accurate modeling. Virtual platooning introduces a higher level of cooperation by organizing vehicles into a logical platoon before physical merging occurs. This enables early synchronization of speeds, precise position adjustment, and proactive gap creation with the target platoon. The technique is mostly used in on-ramp joining scenarios, where the merging vehicle is virtually integrated into the main-road platoon in advance, as demonstrated in \cite{li2024} and \cite{huang2019}.

Recently, reinforcement learning (RL) algorithms have been increasingly adopted to tackle the complexities of platoon-joining tasks. Early foundational work primarily focused on single-agent settings. In the value-based RL domain, \cite{wang2024dqn} introduced a deep Q-network (DQN) solution for platoon merging within a discrete-cell road environment. The results show that the proposed DQN approach requires less merging travel time and fewer vehicle lane-change times than the rule-based approach. Concurrently, policy-based RL methods were explored in \cite{ye2020}, which applied proximal policy optimization (PPO) to automated lane-changing maneuvers and achieved a 95\% success rate in dense traffic. To address the inherently multi-agent nature of platoon coordination, subsequent research extended RL frameworks to handle inter-agent competition and cooperation. A series of multi-agent RL algorithms have been applied to CAV platoon formation, as demonstrated across studies \cite{kolat2024}, \cite{zhou2022}, \cite{zhang2022drl}, \cite{wang2024cav}, \cite{zhou2021}. However, a critical step toward real-world deployment lies in navigating mixed-traffic environments, where CAVs and human-driven vehicles (HDVs) coexist. In this context, \cite{shi2025} proposed a two-stage framework: the first stage enumerates all feasible platoon formations given a specific vehicle sequence, while the second stage employs a multi-agent policy to guide each vehicle—both automated and human-driven—into its designated position according to the formation plan. To further accelerate convergence, hybrid approaches have also been investigated, such as combining genetic algorithms with deep RL for smart platooning \cite{prathiba2021}. Beyond formation planning and convergence speed, collision during the joining maneuver remains a persistent and fundamental challenge. The literature reveals two predominant safety strategies. The first appends an independent safety controller—an external, rule-based module—to override potentially hazardous decisions and ensure collision-free execution \cite{krasowski2020}. The second incorporates internal risky-joining penalties directly within the reward function, thereby discouraging unsafe behaviors intrinsically during the training process \cite{ye2020}. A recent study integrates a control barrier function (CBF) into an RL framework to guarantee a safe joining process in the formation of mixed platoons (including CAVs and HDVs), demonstrating promising results for longitudinal and lateral control \cite{Zhou2024}.

Despite these advances, existing studies indicate that most frameworks and simulations rely on oversimplified or abstracted environments that fail to capture realistic microscopic traffic dynamics. The transition from mixed-traffic lanes to dedicated CAV lanes—a highly practical scenario—has received limited systematic investigation, which was deemed efficient for flow organization with mixed traffic \cite{Kim2023}. Most critically, while safety modules (external shields) and penalty-based rewards (internal costs) have been individually explored, there is a lack of comprehensive comparative analysis that evaluates their distinct roles, trade-offs, and combined effectiveness under diverse and dynamic traffic conditions. To bridge these gaps, this paper proposes a generic RL-based modeling framework built within the SUMO simulation environment to systematically study CAV platoon joining maneuvers. 

Our main contributions therefore are threefold: 1) developing a system framework that enables CAVs to join a platoon when operating from a mixed-traffic lane to a dedicated CAV lane on the highway; 2) implementing and comparing three distinct types of deep RL algorithms to learn optimal joining policies; and 3) conducting comprehensive performance analyses under diverse traffic conditions, with a specific emphasis on assessing the roles of risky penalties during training versus the independent safety shield for collision avoidance, thereby providing critical insights into the design of robust and reliable platoon-joining controllers.

The rest of the paper is organized as follows. We introduce the fundamentals of RL and typical deep RL algorithms in Section 2. The detailed methodology, including the simulation system architecture, platooning task modeling, and algorithm implementation, is presented in Section 3. Afterwards, we present the simulation experiments and results analysis. The conclusion is drawn in the last section. 

\section{Background Knowledge}
\label{background}
\subsection{Fundamentals of RL}
\label{sub_reinfor}
RL is designed for an agent to learn by interacting with the environment. The agent learns how to make decisions by taking actions and receiving feedback from the environment. Specifically, the decision-making process is modeled as a Markov decision process (MDP), defined by the tuple $(\mathcal{S}, \mathcal{A}, P, r, \gamma)$, where $\mathcal{S}$ and $\mathcal{A}$ denote the state and action spaces, respectively, $P(s'|s, a)$ represents the state transition probability, $r(s, a, s')$ is the reward function, and $\gamma\in \left( 0, 1 \right]$ is the discount factor for future reward. At each time step $t$, the agent selects action $a_t$ based on the current state of the environment $s_t$. The environment then responds by transitioning to a new state $s_{t+1} \sim P(\cdot |s_{t}, a_{t})$ and returns a reward $ r_{t+1} = r(s_{t}, a_{t}, s_{t+1})$. The goal of RL is to learn a policy $\pi: \mathcal{S}\to\mathcal{A}$ in the form of a condition probability $\pi(a_t|s_t)$ to maximize the expected cumulative discounted reward in infinite or over a finite horizon $T$, defined as $J(\pi) =\mathbb{E}_{\pi}\left[ \sum_{t=0}^{T-1}\gamma^{t}r_{t+1} \right]$. Q-learning is a classical RL method based on the Bellman Equation and seeks to maximize the objective $J(\pi)$ by updating the estimate of $Q(s,a)$ iteratively.

\subsection{DQN and DDQN}
\label{sub_deep_reinfor}
When deep neural networks are used to approximate the value function (e.g., the Q-function) or the policy, the approach is known as deep RL. As one of the deep RL algorithms, DQN uses the same Q-value maximization for both action selection and evaluation, although the online and target networks are maintained separately. The target value function is given by, 
\begin{equation}
y^\text{DQN} = r + \gamma\underset{a'}{\text{max}}Q(s', a';\theta^-)
\end{equation}
where $\theta^-$ are the weights of the target network. The goal is to find $\theta$ in the Q-network that estimates the best $Q$ function, i.e., $Q(s,a; \theta) \approx Q^{*}(s,a)$ by minimizing the following loss function:
\begin{equation}
\label{eq:loss}
 \mathcal{L}(\theta) = \mathbb{E}\left[(y^\text{DQN}-Q(s, a; \theta))^2\right]
\end{equation}
where $\theta$ can be optimized using gradiant decent and $\theta^-$ is periodically cloned from $\theta$ (e.g., a soft update strategy) to maintain stability. 

Due to noise in the estimated $Q(s,a; \theta)$, the max operator can lead to a systematic overestimation bias. Therefore, DDQN \cite{mnih2013} is proposed to use two different networks (with the same architecture), where the online network picks the best action, and the target network evaluates its value, as shown in Eq.(\ref{eq:ddqn}). 
\begin{equation}
\label{eq:ddqn}
y^\text{DDQN} = r + \gamma Q(s',\text{arg}\underset{a'}{\text{max}}Q(s', a';\theta);\theta^-)
\end{equation}
where argmax uses the online weight $\theta$ and the Q value uses the target weights $\theta^-$. The loss function in Eq. (\ref{eq:loss}) then uses $y^\text{DDQN}$ instead of $y^\text{DQN}$.

Normally, the $\epsilon$-greedy policy is used for action selection. The best action is chosen with probability $1-\epsilon$ and the random action selection with probability $\epsilon$. Specifically, we adopt the time-dependent $\epsilon(t)$ determined by an exponential decay function as shown in Eq. (\ref{eq: epsilon}). This $\epsilon(t) = \epsilon_{\min}$ is satisfied after a large number of simulation steps of $\epsilon_{\text{decay}}$, which means the best actions are mostly chosen in the final convergence.

\begin{equation}
\label{eq: epsilon}
\epsilon(t) = \exp \left( 
    \log(\epsilon_{\text{start}}) 
    + \frac{t}{\epsilon_{\text{decay}}} 
      \big( \log(\epsilon_{\min}) - \log(\epsilon_{\text{start}}) \big) 
\right)
\end{equation}

For both algorithms, a replay buffer is used, in which a batch of stored experience quadruplets (state, action, reward, new state) is randomly sampled for $\theta$ updates.
\subsection{PPO}
PPO does not use a Q-target network. It updates its policy using an advantage function in Eq.(\ref{eq:ppo}) (usually computed via GAE - Generalized Advantage Estimation) to tell the agent how much better an action is compared to the average.
\begin{equation}
\label{eq:ppo}
\hat{A_t} = \delta_{t} + (\gamma\lambda)\delta_{t+1} + ... + (\gamma\lambda)^{T-t+1}\delta_{T-1}
\end{equation}
where $\delta_{t}=r_t + \gamma V(s_{t+1}) - V(s_t)$ and $\lambda$ is the GAE parameter. The objective in PPO is to maximize a surrogate objective function (see Eq.(\ref{eq:surrogate})) by using importance sampling to safely reuse old data to update the neural network's weights $\theta$. 
\begin{equation}
\label{eq:surrogate}
L^\text{PPO}(\theta)=\mathbb{E}_t \left[\text{min}(r_t(\theta)\hat{A_t}, \text{clip}(r_t(\theta),1-\epsilon,1+\epsilon)\hat{A_t}) \right]
\end{equation}
The related PPO algorithm can be seen in \cite{ye2020}.

\section{Methodology}
\label{metho}
\subsection{System architecture}
\begin{figure}[t]
    \centering
    \includegraphics[width=\linewidth]{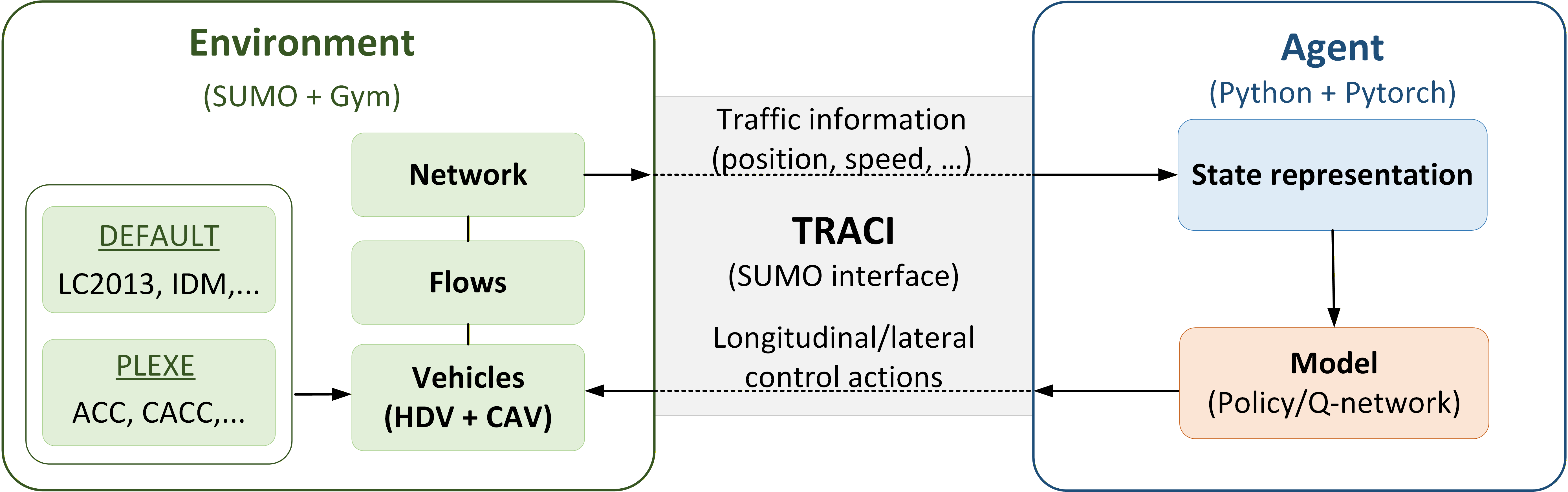} 
    \caption{System architecture for the RL-based agent modeling and simulation}
    \label{fig:system}
\end{figure}

To realize the agent-based learning process in a traffic dynamic simulation environment, we illustrate the built system architecture in Fig. \ref{fig:system}.

The system consists of two main components: the environment and the agent. For the environment, two-lane highway traffic is modeled in SUMO, along with vehicle characteristics and traffic load variations over time. In the network configuration, the inner lane is dedicated exclusively to CAV platoons, and the outer lane is a mixed lane that includes both HDVs and CAVs. The speed limits of these two lanes can be different. Besides the embedded longitudinal and lateral control models in SUMO for HDVs, the environment includes a CAV control module -- PLEXE \cite{segata2022}, which is a cooperative driving framework extended in SUMO to enable automated driving behavior simulation, such as ACC and CACC. To facilitate communication with the learning agent, the environment is encapsulated as a Gym environment, which is a standard open-source library used for many RL applications. The current environmental information around the ego CAV is transmitted via the SUMO Traci API to construct the state representation, which is used as input for the agent-based learning model. The model predicts values of possible actions in terms of longitudinal or lateral driving control, and the selected action is sent back to the ego CAV for maneuver execution. Notably, the built system architecture is supposed to be transferable for other agent-based traffic control problems, such as traffic signal control \cite{Ducrocq2023} and on-ramp merging \cite{Dinneweth2026}.   

\subsection{CAV platoon formation problem}
\label{sec:workflow}
\begin{figure}[t!]
    \centering
    \includegraphics[width=0.8\linewidth]{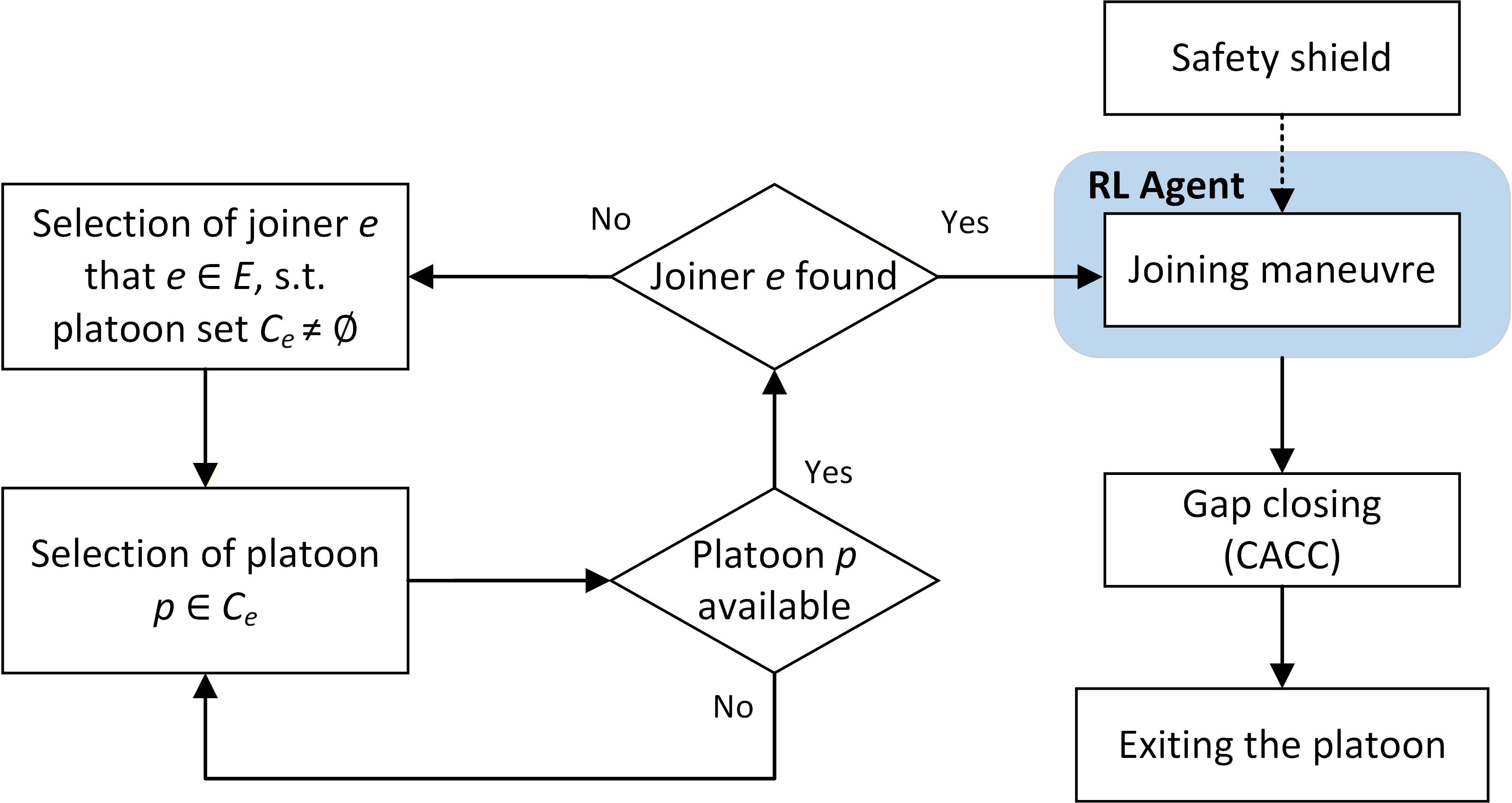} 
    \caption{Workflow of platooning control}
    \label{fig:workflow}
\end{figure}

In this study, we assume that the joining CAV enters at the rear of the target platoon and leaves from the rear position. All members in the platoon from the head to the last are organized in increasing order according to their destinations (i.e., exits of the highway). This strategy reduces the need for internal reordering and mitigates traffic disturbances. Focusing on our CAV platoon formation, the control procedure consists of three phases: 1) CAV joiner and platoon selection, 2) joining maneuver, and 3) gap closing. Fig. \ref{fig:workflow} illustrates the workflow of platoon formation. Regarding the life cycle of the platoon, an additional process for exiting platoons should be integrated. This process is not in the scope of this paper. The above three main phases for the platoon formation are introduced below.

\paragraph{Joiner and target platoon selection.}  A CAV is randomly selected from the set of joiner candidates on the mixed-traffic lane (i.e., outer lane). The set of joiner candidates consists of CAVs that must have the possibility to choose one of the existing platoons within their perception range. Here, the perception range is not the same as the communication zone, in which joining requests can be reached as far as possible. We define the perception range as follows: if the joiner is at position $x$ meters (m), the last member of the target platoon should be located in the range of $[x - \delta_1, x + \delta_2 ]$. $\delta_1$ is the distance behind the joiner vehicle to ensure the maximum platooning size, and it must be less than the radius of the communication zone $R$. $\delta_2$ is a short distance ahead of the joiner vehicle for following the target platoon in a short time. In our study, we set $\delta_1 = 150$ m, $\delta_2 = 5$ m and $R =200$ m. When platoons receive the joining request from the selected CAV joiner, they may accept or refuse the joining request, subject to their availability. The platoon's availability depends on two conditions: 1) no ongoing joining/splitting process by another CAV, and 2) satisfying the ordered destinations regarding the existing CAV members in the platoon. Under these constraints, a joiner vehicle will be finally confirmed to join one of the available platoons in its neighborhood. If there is no platoon in the zone, a CAV can move to the inner lane under the default lane-change model to form an initial platoon containing only one vehicle.

\paragraph{Joining maneuver.} Once the CAV joiner and the target platoon are selected, the joining maneuver (either lane change or speed adjustment) starts under control by the agent learning model. This phase mainly focuses on implementing the deep RL algorithms (i.e., DQN, Double DQN, and PPO). The CAV joiner will learn from the surrounding traffic conditions (i.e., state), intending to join the target platoon by executing an appropriate driving maneuver (i.e., action), through either adjusting longitudinal speed or switching to the inner lane, under the assessment of the cumulative reward. The configuration of each component of the deep RL algorithms will be detailed in Section \ref{RL_agent_modeling}. To avoid collisions, related safety constraints such as time-to-collision (TTC) can be applied during the algorithm training or before the execution of lane changes. We will initially take TTC-related collision risks into account during the model training for the purpose of assessing the learning performance. A TTC-related safety shield will be considered for a comparison as needed.    
\paragraph{Gap closing.}
After the ego vehicle $V_e$ arrives at the target platoon lane, it should approach the target platoon and minimize the coordinated distance $d_{crd}$ (e.g., 5 m) to its predecessor. To achieve this, we use the embedded PLEXE package for the CACC implementation, which allows the ego vehicle to synchronize its behavior to the target platoon members and complete its formation. Since the inner lane is dedicated exclusively to platoons composed of CAVs and the ego vehicle is correctly positioned after the lane-change maneuver, this simple mechanism is sufficient to maintain the desired headway among platoon members.

\subsection{RL for CAV agent modeling}
\label{RL_agent_modeling}
\begin{figure}[t]
\centering
\includegraphics[width=0.8\linewidth]{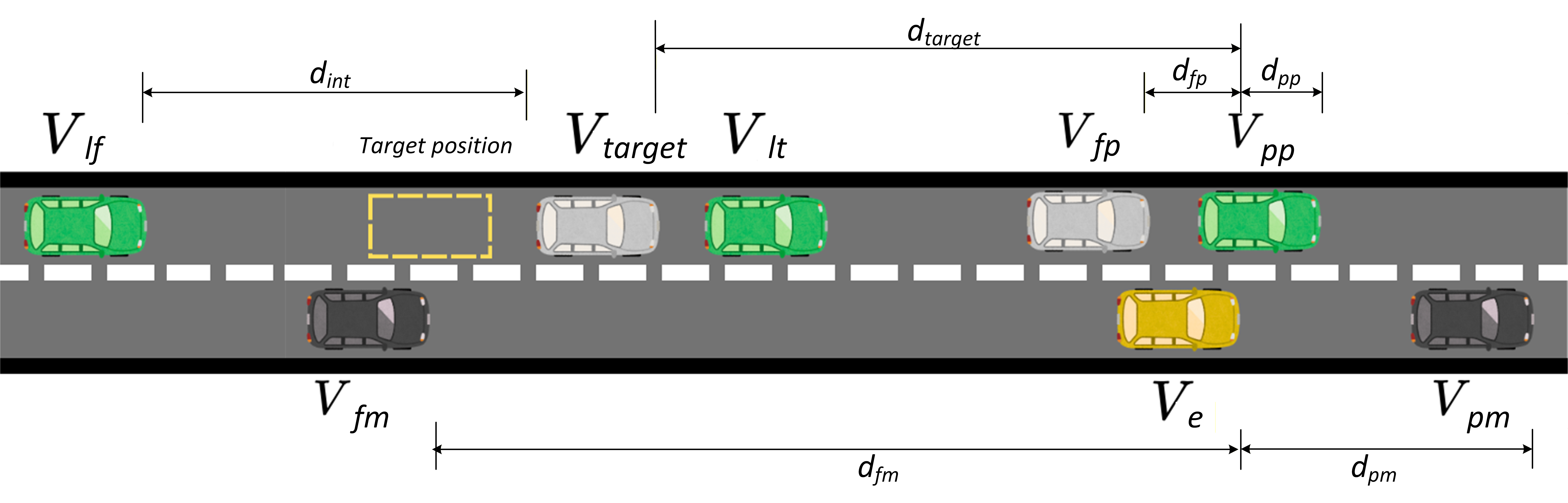}
\caption{Ego vehicle and surrounding vehicles}
\label{fig:sur_veh}
\end{figure}

\paragraph{State representation.} For a CAV joining a platoon at the right moment, it should have a vision of the surrounding vehicles. This is crucial for the ego vehicle to adapt speed and know when to change lanes without causing collisions.  Fig. \ref{fig:sur_veh} illustrates the CAV joiner and its surrounding vehicles as follows: 
\begin{itemize}
\item $V_e$: the CAV ego vehicle, i.e., the joiner.
\item $V_{pm}$: the predecessor of the ego vehicle on the mixed lane $m$.
\item $V_{fm}$: the follower of the ego vehicle on the mixed lane $m$.
\item $V_{pp}$: the predecessor of the ego vehicle on the platoon lane $p$.
\item $V_{fp}$: the follower of the ego vehicle on the platoon lane $p$.
\item $V_{lt}$: the leader of the target platoon.
\item $V_{target}$: the last member of the target platoon. It can be the same as $V_{lt}$ if the platoon initially consists of a single vehicle.
\item $V_{lf}$: the leader of the following platoon.
\end{itemize}

Here, the state is a vector of $15$ variables given by Eq. (\ref{eq:state}).
\begin{equation}
\label{eq:state}
\begin{aligned}
& s = \{\, &v_{e}, v_{pm}, v_{fm}, v_{pp}, v_{fp}, v_{target}, v_{lf},
d_{pm}, d_{fm}, d_{pp}, d_{fp}, d_{target}, d_{int}, l_e, l_p \,\}
\end{aligned}
\end{equation}
where $v_{e}, v_{pm}, v_{fm}, v_{pp}, v_{fp}, v_{target}$, and $v_{lf}$ are respectively the speeds of the ego vehicle $V_e$, its predecessor and follower vehicles on the mixed lane (i.e, $V_{fm} ,V_{fm}$), its predecessor and follower vehicles on the platoon lane (i.e., $V_{pp}, V_{fp}$), and the predecessor $V_{lt}$ and follower $V_{lf}$ of the target joining position; $d_{pm},\allowbreak d_{fm},\allowbreak d_{pp}$, and $d_{fp}$ are the absolute distances between the ego vehicle and the predecessor/follower vehicles on the two lanes; $d_{target}$ is the directional distance between the ego vehicle and the last vehicle in the target platoon (notice that $d_{target}$ is negative when $V_{target}$ is behind of $V_e$, otherwise.);  $ d_{int} $ is the absolute value of the distance between the rear of $V_{target}$ and the head of $V_{lf}$. Except for $ d_{int} $, for simplification, all distances are calculated as head distances between the ego vehicle and the related surrounding vehicles. $l_e$ and $l_p$ are the indices of the ego vehicle's located lane and the platoon lane. 
We assume that related vehicle information can be acquired in time within the communication zone. In a low-density traffic situation, the predecessor and follower vehicles may be partially missing. We adopt the default limit speed on the lane and the communication range $R$ as substitutes, e.g.,$v_{pm}$ = 100 km/h and  $d_{pm}$ = 200 m when no vehicle is ahead of $V_e$.     

\paragraph{Action space.} Our action space consists of six discrete actions, including the lane change action $a_0$ with a current speed and five actions for speed adjustements without lane changes, which are defined as follows: $a_1$ = $2\,\mathrm{m/s^2}$, $a_2$ = $1\,\mathrm{m/s^2}$, $a_3$ = $0\,\mathrm{m/s^2}$, $a_4$ = $-1\,\mathrm{m/s^2}$, and $a_5$ = $-2\,\mathrm{m/s^2}$.
To simulate an action in a reasonable implementation, we set a decision-action interval $\Delta t$ consisting of five simulation steps (i.e., a total of 0.5 s due to the simulation time step of 0.1 s), meaning that once an action is chosen, the next action can only be taken after five simulation steps. The decision-action interval should ensure the time delay from the decision-making process and the communication between vehicles.

\paragraph{Reward function.} To achieve an efficient and safe joining maneuver, we propose a reward function with sub-rewards in Eq. (\ref{eq: reward}), implying a bonus of a successful joining and penalties in other cases such as a failure joining, a joining with a collision, a delayed joining, and unstable joinings (either being risky as too close to the last member of the target platoon or being inefficient as farway behind). The reward formulation is presented as follows:
\begin{equation}
\label{eq: reward}
\begin{array}{lcl}
\displaystyle r = \alpha_s(r_s + r_r + r_u) + \alpha_f\cdot r_f + \alpha_c\cdot r_c  + \alpha_k(r_r + r_j)
\end{array}
\end{equation}
where: 
\begin{itemize}
    \item $\alpha_s$, $\alpha_f$, $\alpha_c$, and $\alpha_k$ are dummy variables. For each variable, a value of 1 indicates that the corresponding status (success, failure, collision, or lane keeping) occurs, while 0 indicates otherwise.
    \item $r_s$ is a success bonus. A successful joining maneuver leads the ego vehicle $V_e$ to the target position behind the predecessor $V_{target}$ in the target platoon. The success bonus is set to $100$.
    \item $r_r$ is the risky maneuver penalty, although the joiner gets a successful join. It is related to two alternative penalties for a risky situation of either changing to the platoon lane ($\alpha_s = 1$) or staying on the mixed lane with speed adjustments ($\alpha_k = 1$), according to the measures of rear and front TTC. We set it as:
    \begin{equation}
    \label{eq: risky_penalty}
    \begin{array}{lcl}
    \displaystyle r_r =
    \begin{cases}
        -100, & if\ \text{min}(TTC_{(V_e, V_{lf})}, TTC_{(V_e, V_{target})}) < 2  \\
         & and\ \alpha_s = 1, \\[6pt]
        -50, & if\ \text{min}(TTC_{(V_e, V_{fm})}, TTC_{(V_e,V_{pm})}) < 1 \\
        & and\ \alpha_k = 1, \\[6pt]
        0, & otherwise.
    \end{cases}
    \end{array}
    \end{equation}
    
    \item $r_u$ is the unstable maneuver penalty. This is related to the distance (either too close or too far) to the target platoon after joining. We consider the range of safe positions between 10 m and 20 m without any penalty. Eq.(\ref{eq: unstable_penalty}) represents how this sub-reward is formulated:
    \begin{equation}
    \label{eq: unstable_penalty}
    \begin{array}{lcl}
    \displaystyle r_u(d) =
    \begin{cases}
        -2  (10 - d), & if\  d < 10, \\[6pt]
        0, & if\ 10 \leq d \leq 20, \\[6pt]
        -0.5 (d - 20), & otherwise.
    \end{cases}
    \end{array}
    \end{equation}
    where $d$ is the immediate distance between the joiner $V_e$ and the vehicle $V_{target}$ after a joining maneuver (i.e., only completing a lane change before the gap closing).
    \item $r_f$ is the failure penalty. Any joining maneuver by which the joiner $V_e$ is not placed after the predecessor $V_{target}$ is considered a failure. A failure also occurs when the joiner exits the network before performing the joining maneuver. We set the failure penalty to $-50$.
    \item $r_c$ is the collision penalty. To avoid collisions, we set a heavy penalty of -10,000 in a collision case that occurs either on the platoon lane or on the mixed lane.
    \item $r_j$ is the jerk penalty. This penalty relates to excessive longitudinal acceleration and jerk when actions are taken to keep the current lane with the speed adjustment. We penalize the jerk if it satisfies $\left| \dot{a} \right| > 4\,\mathrm{m/s^3}$ (when $\alpha_k = 1$), reflecting an uncomfortable driving experience. In our study, the penalty is set as:
    
    \begin{equation}
    \label{eq: jerk_penalty}
    \begin{array}{lcl}
    \displaystyle r_j =
    \begin{cases}
        -\left| \dot{a} \right|, & if \left| \dot{a} \right| > 4, \\[6pt]
        0, & otherwise.
    \end{cases}
    \end{array}
    \end{equation}

\end{itemize}

\section{Simulation and Results}
\subsection{Experimental setup}
To build a mixed CAV traffic environment in SUMO, we adopt a CAV penetration ratio of 50\% on the mixed lane coexisting with HDVs and set 100\% CAVs on the platoon lane. The default car-following model (IDM) and lane-change model (LC2013) are applied only for HDVs. Thanks to the package PLEXE embedded in SUMO, for CAVs not in platoons (i.e., on the mixed lane) and platoon leaders (i.e., on the platoon lane), the ACC car-following model is adopted; on the platoon lane, the CAV ego vehicle adopts the CACC car-following model to catch up to the target platoon. Once a CAV is selected as a platoon joiner, its ACC or CACC is deactivated, and lane changes and speed adjustments are governed by the deep RL algorithms. Table \ref{tab:parameter_DQN_DDQN} and Table \ref{tab:parameter_PPO} show the hyperparameter settings for training/testing the value-based models of DQN and DDQN, and the policy-based PPO model, respectively. Table \ref{tab:parameter_scenarios} gives the configuration for simulating traffic scenarios. All experiments were performed on a server featuring a 2.4 GHz CPU with 64 processors and 512 GB of RAM.

\subsection{Model training performance}
Before training, a warm-up phase of 1,200 simulation steps is launched in SUMO to allow vehicles to spread out in the network, and platoons are also initialized during the phase. Specifically, in our configuration, we set \textit{event} as the unit of executed joining status (i.e., success, failure, or collision). Each training \textit{episode} consists of 1,000 decision-making steps, thus resulting in fewer than 1,000 events. That is to say, one event could occur within multiple decision steps, as a selected joiner can either stay on its current lane due to an adopted action for just speed adjustment, or give up its platoon joining if potential collision risks are detected (see safety shield tests in Section \ref{subsec_safety_shield}). We train the model by $10^6$ decision steps with incremental levels of random traffic. The training time costs are: 46.8 hours for the DQN, 18.9 hours for the DDQN, and 12.1 hours for the PPO. 

\begin{figure}
\centering
\includegraphics[width=0.9\linewidth]{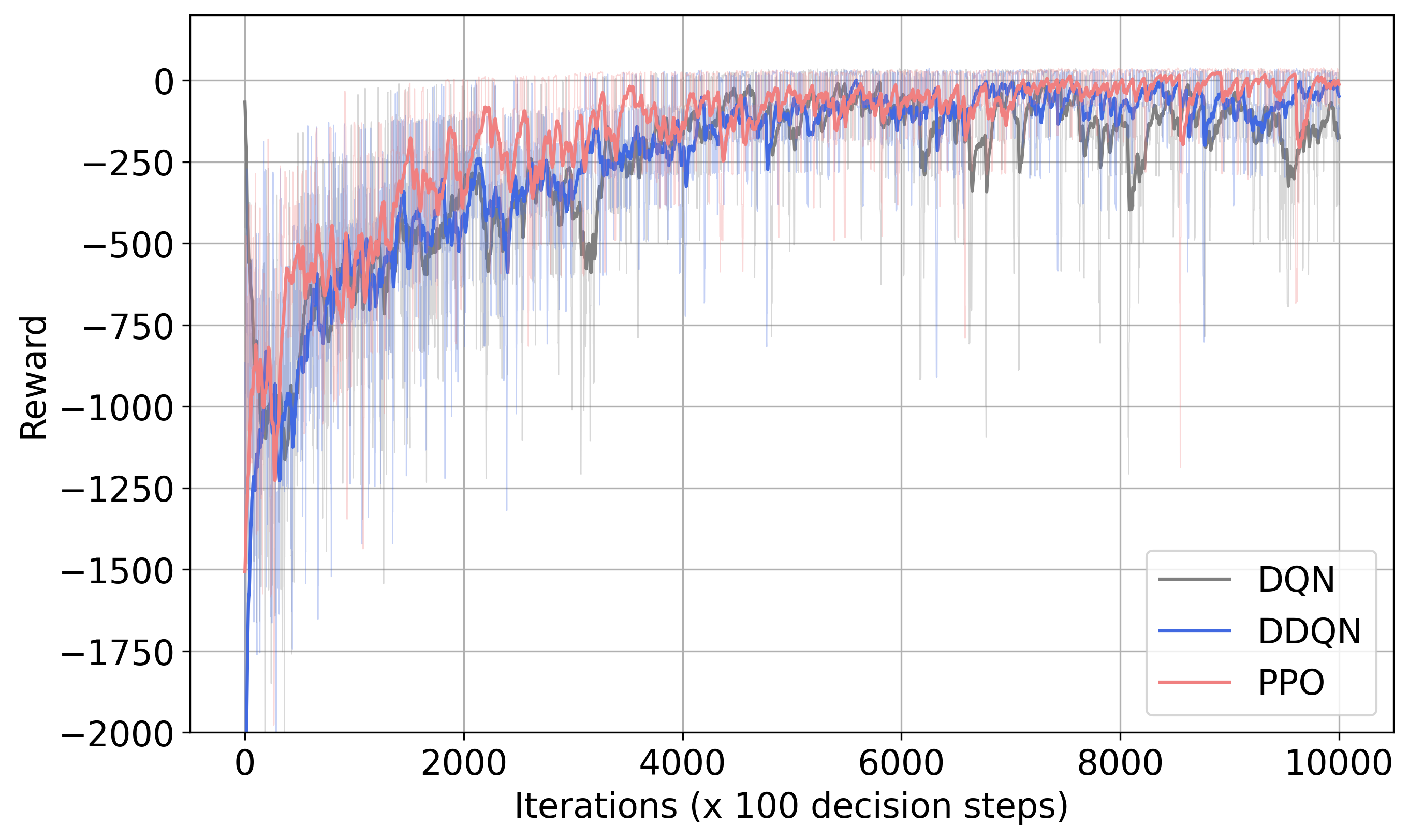}
\caption{Reward evolutions}
\label{fig:rew}
\end{figure}

The generated average rewards per recording log iteration of the last 100 decision steps are shown in Fig. \ref{fig:rew}. To make the comparison observable, we smoothed the rewards to represent their evolution, which increases over iterations. Overall, PPO outperforms the other two algorithms, especially during the two periods before 4,000 iterations and after 8,000 iterations, which almost correspond to low and high traffic load scenarios. Compared to DQN, the PPO and DDQN algorithms have relatively more stable and larger mean rewards ($<$-50) for the last periods. Small negative rewards indicate that collisions, failures, or risky joining occur very occasionally.

\begin{figure}
\centering

\begin{subfigure}{0.44\textwidth}
    \centering
    \includegraphics[width=\linewidth]{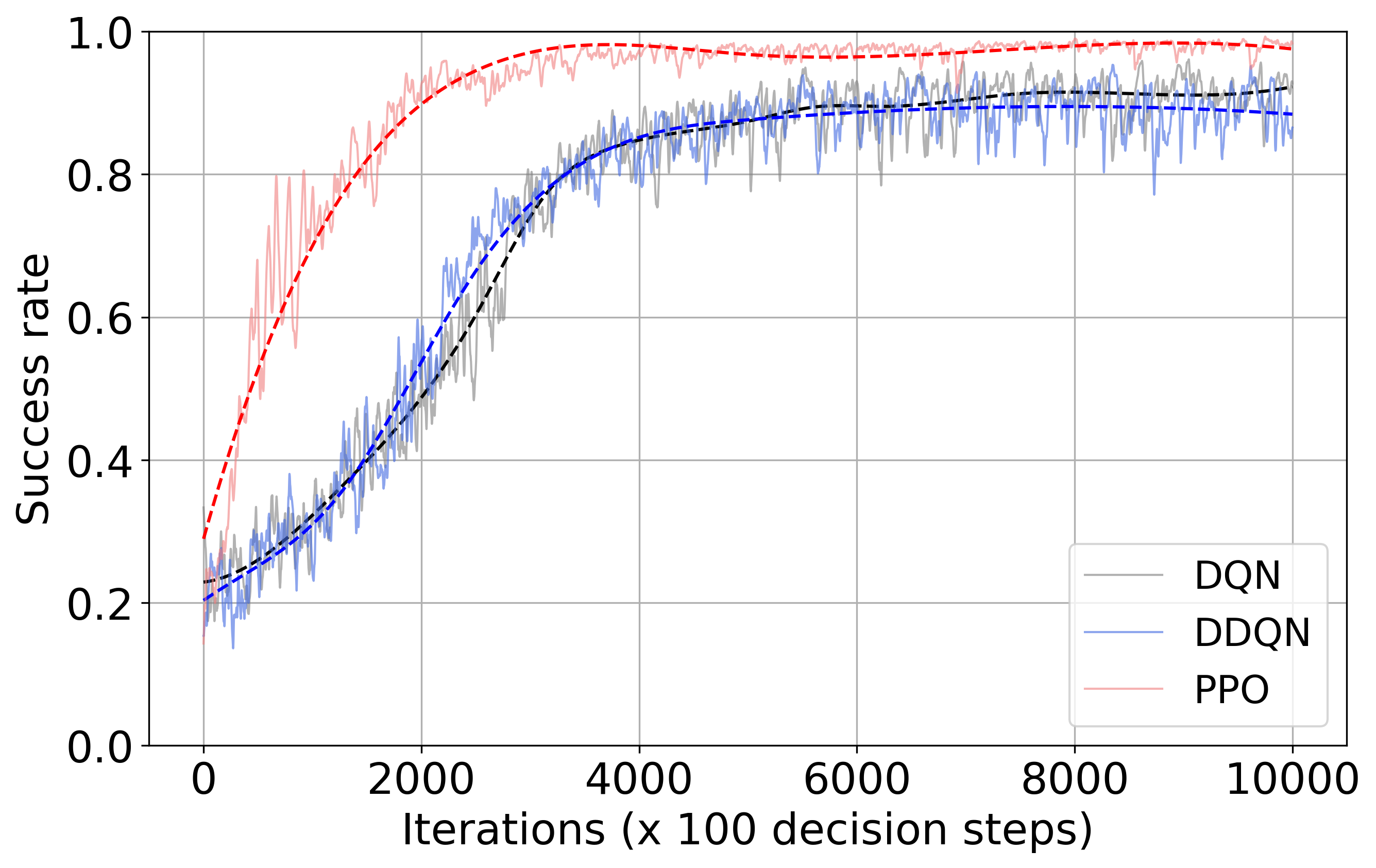}
    \caption{Success rate}
\end{subfigure}
\hspace{0.01\linewidth}
\begin{subfigure}{0.44\textwidth}
    \centering
    \includegraphics[width=\linewidth]{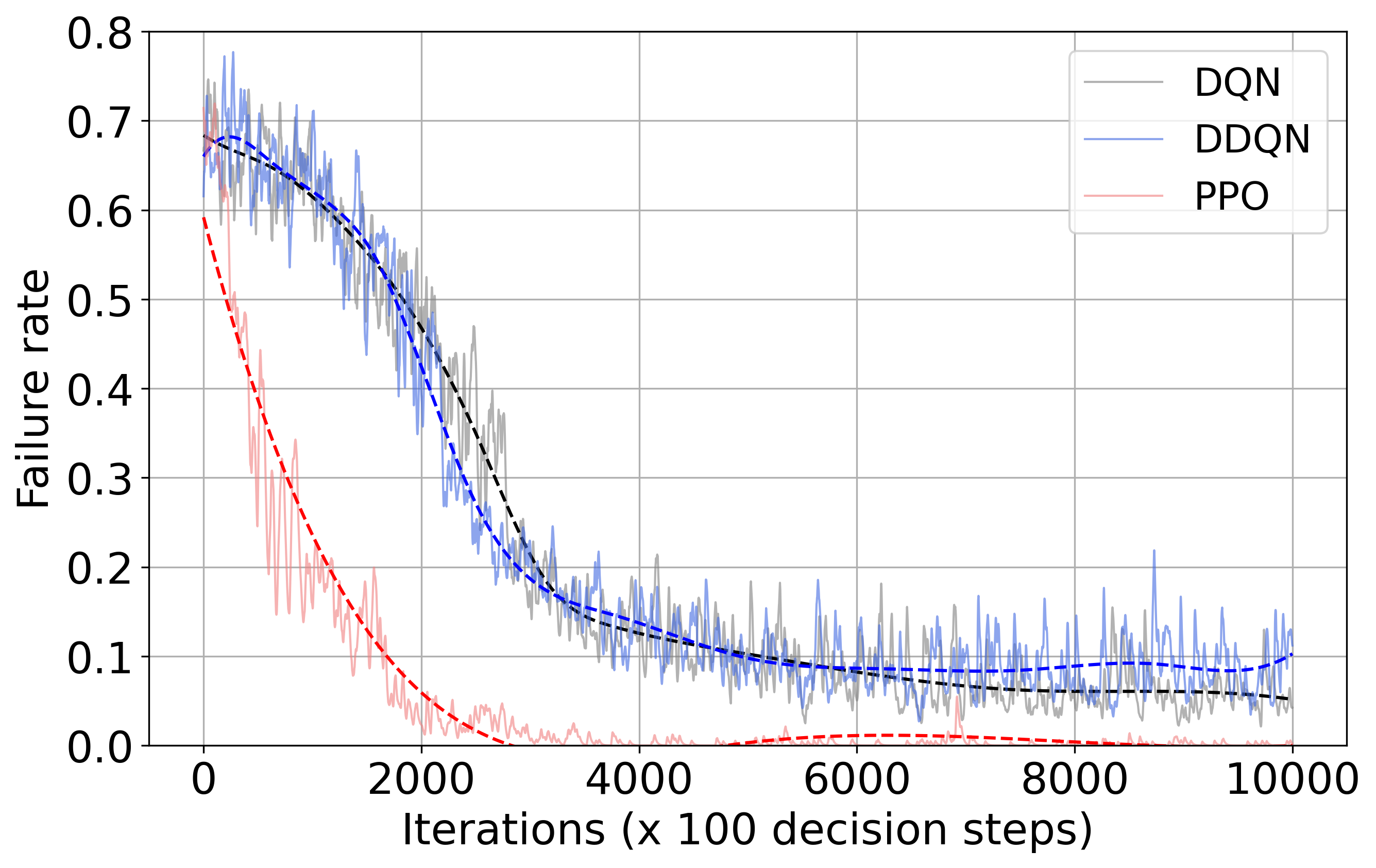}
    \caption{Failure rate}
\end{subfigure}
\vspace{0.5em}
\begin{subfigure}{0.44\textwidth}
    \centering
    \includegraphics[width=\linewidth]{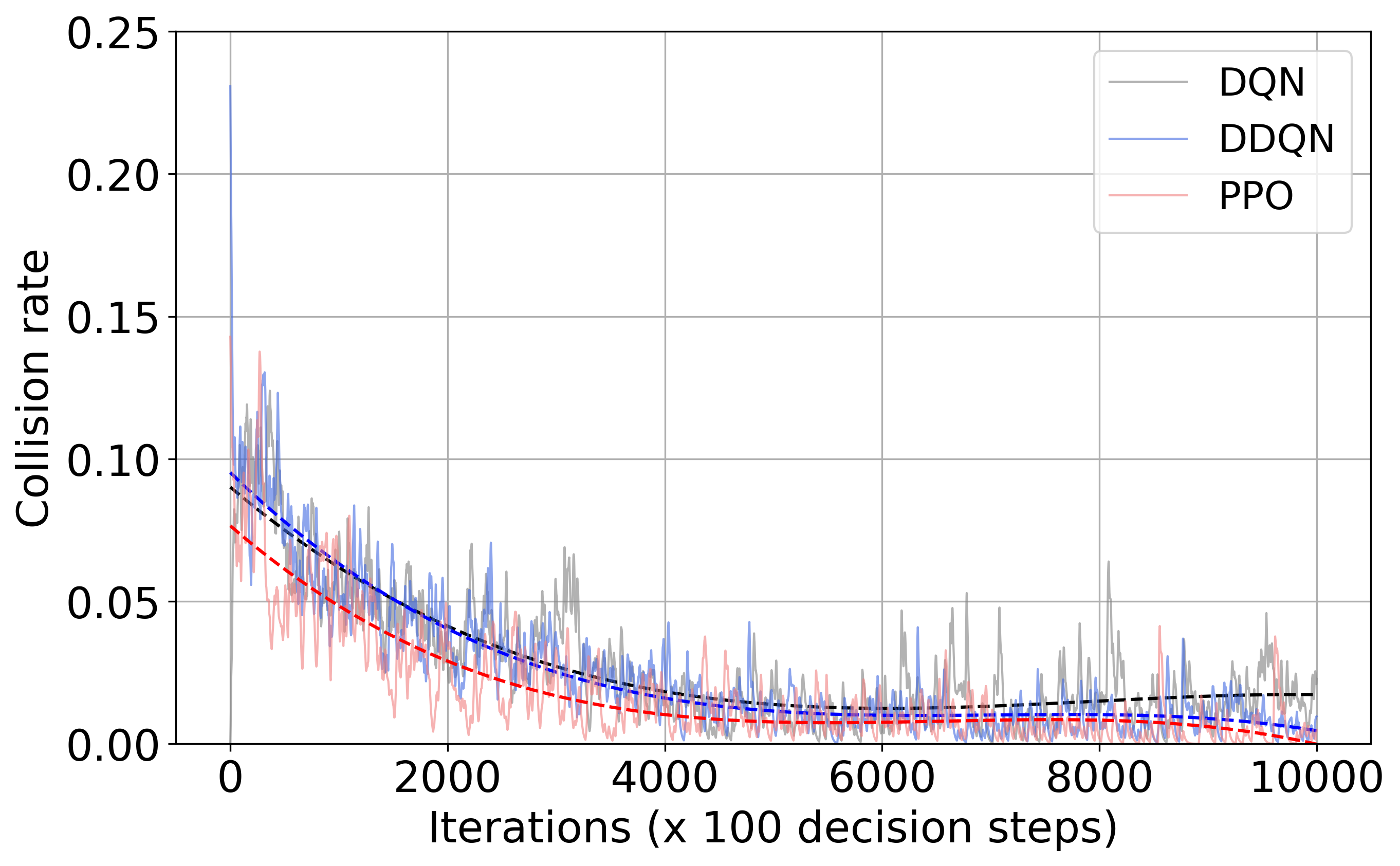}
    \caption{Collision rate}
\end{subfigure}
\hspace{0.01\linewidth}
\begin{subfigure}{0.44\textwidth}
    \centering
    \includegraphics[width=\linewidth]{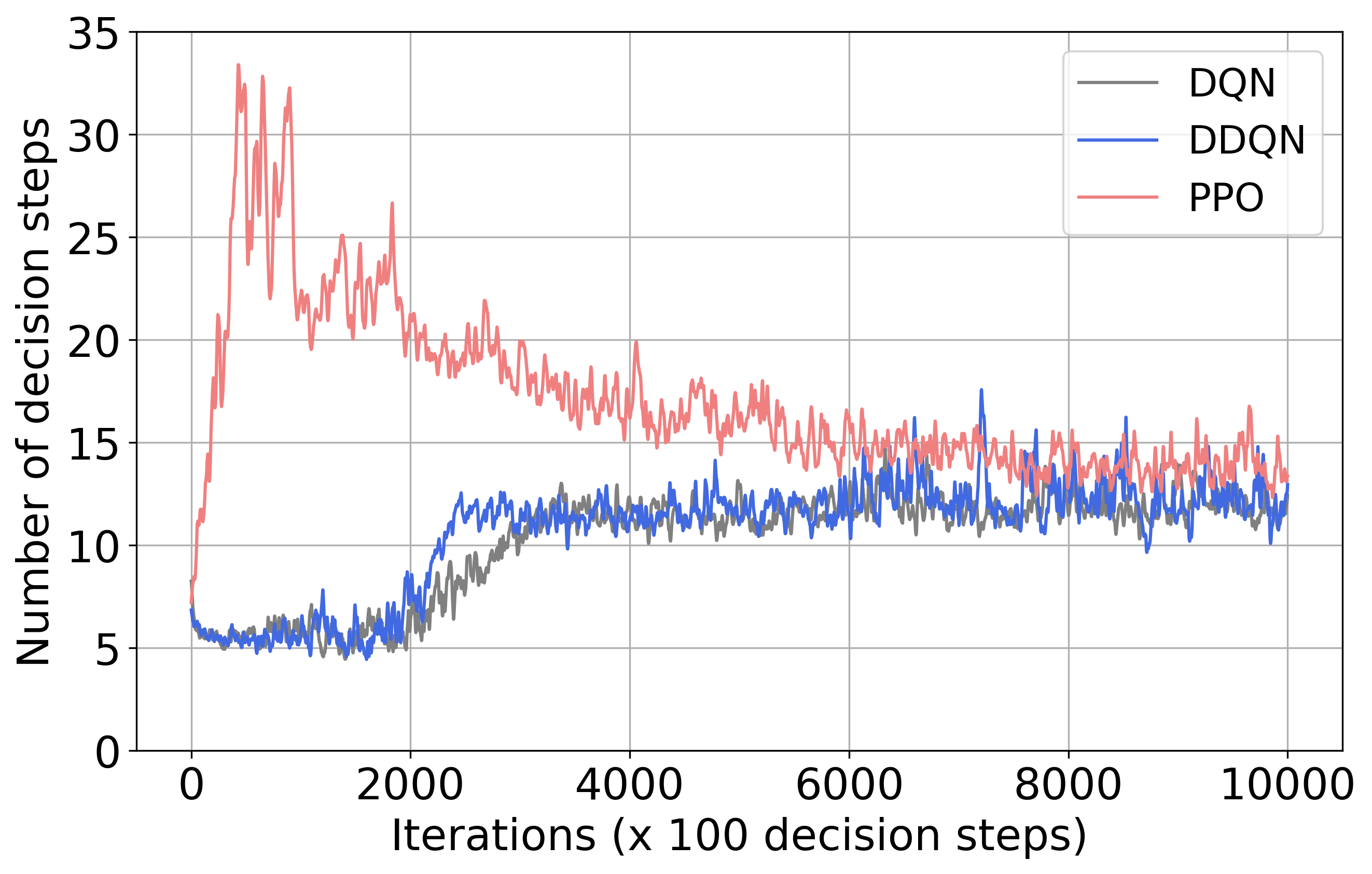}
    \caption{Decision steps per event}
\end{subfigure}
\vspace{0.5em}
\begin{subfigure}{0.44\textwidth}
    \centering
    \includegraphics[width=\linewidth]{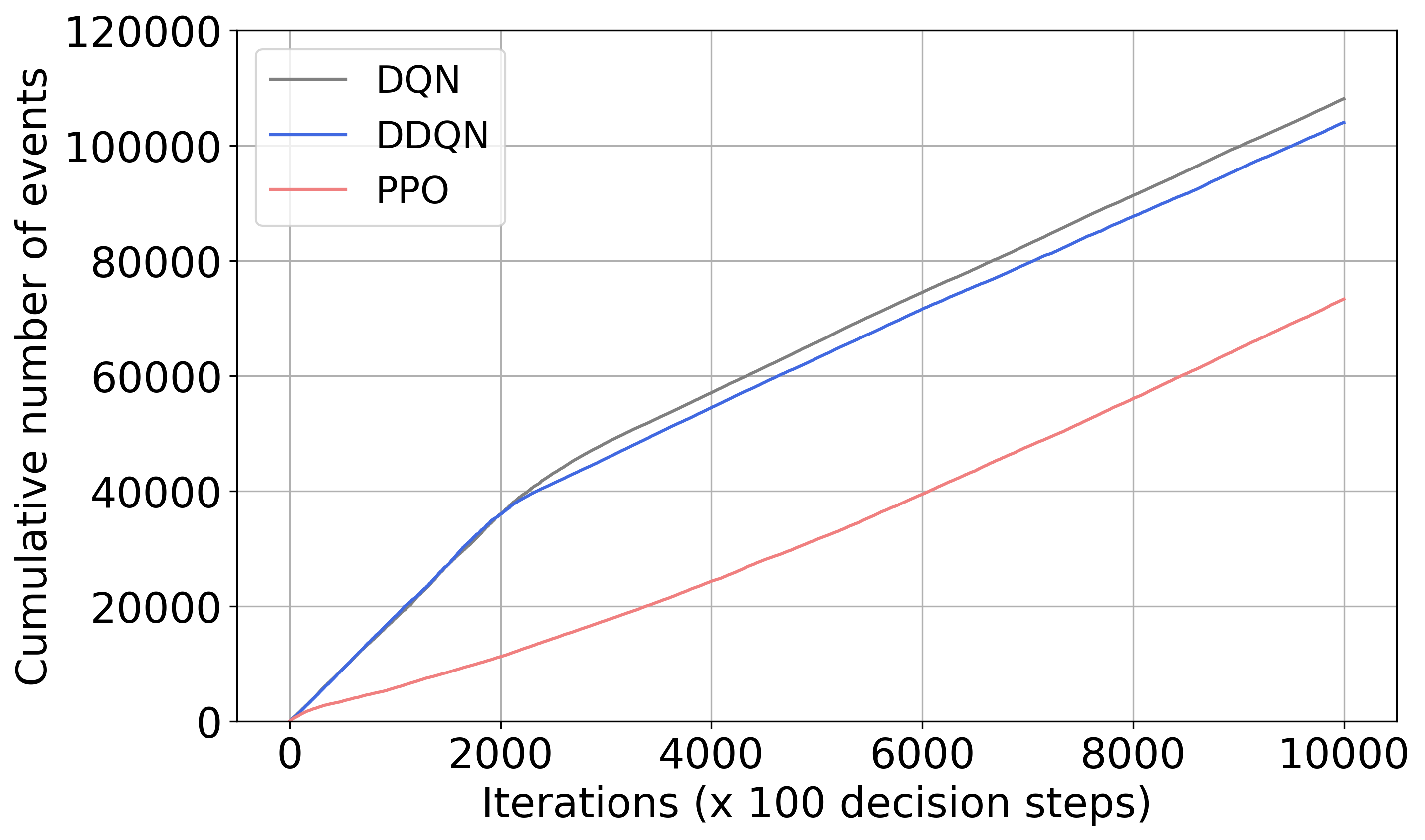}
    \caption{Events}
\end{subfigure}
\caption{Comparisons of training performance.}
\label{fig:learning_metrics}
\end{figure}

We present the operational performances of the three algorithms in Fig. \ref{fig:learning_metrics}. The evolutions of success, collision, and failure rates are shown respectively in Fig. \ref{fig:learning_metrics}(a), (b), and (c). By using a smoothing spline fit, they exhibit similar tendencies where the success rates increase, while the failure and collision rates decrease over time. 
Compared to DQN and DDQN, obviously, PPO achieves superior performance, which has not only the fastest convergence (around 3,000 iterations) but also the most efficiency with the highest success rate and lowest failure/collision rates. DDQN has relatively lower collision rates than DQN. On the contrary, its performance in terms of success rates and failure rates is a bit worse than DQN. Fig. \ref{fig:learning_metrics}(d) represents the convergence of decision steps per event that occurred. PPO decreases the number of decision steps for convergence, while DQN and DDQN are opposites. They all fall in the range of 10 $\sim$ 15 decision steps, namely between 5 and 7.5 seconds (as mentioned, one decision step is 0.5 seconds). Among them, PPO seems more conservative for platoon joining by taking a few more steps. This can also be seen in Fig. \ref{fig:learning_metrics}(e), where its cumulative number of events is less than the other two algorithms entirely. From this, it explains well why PPO consumed so much less time to complete the training.  

Regarding the value-based DQN and DDQN algorithms, initially, the agent randomly selects actions to explore the environment with a strong possibility of a high exploration rate $\epsilon$. In cases where the selected vehicle's target position is behind, for instance, any early lane-change manoeuvre will lead to either a collision or a join in the wrong position. This is why, at the beginning of training, the failure rate is very high ($>70\%$), and the collision rate reaches $10\%$. Successful join maneuvers occur occasionally ($20\%$). With $\epsilon$ decreasing exponentially over time, the probability (i.e., $1 - \epsilon$) of selecting a greedy action gets higher. The agent learns to successfully join the platoon, reaching approximately $98\%$, and the failure and collision rates tend to be smaller near zero. We should mention that at the end of training, the success rate still oscillates because we keep a small exploration rate ($\epsilon_\text{min}$ = 0.01) to ensure continuous learning. This means the agent selects random actions with a probability of $1\%$, which may lead to collisions or failures occasionally.

\subsection{Model testing analysis}
\subsubsection{Joining manoeuvre and speed synchronization}
\begin{figure}
    \centering
    \includegraphics[width=0.8\linewidth]{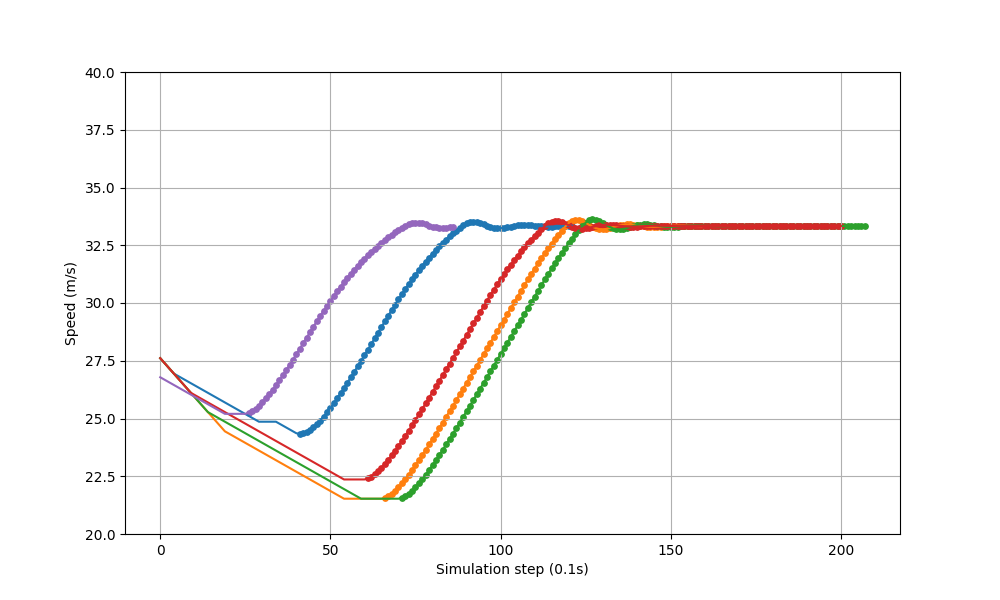}
\caption{Samples of joiners' speed evolutions}
\label{fig:speed}
\end{figure}

We analyze the joining maneuver based on the speed profiles of the joiner samples, as illustrated in Fig. \ref{fig:speed}. The solid lines represent the joiners staying on the mixed lane or starting the lane change (controlled by the deep RL algorithm), and the dotted lines represent ongoing lane changes and catching up with the last members of the platoons (controlled by CACC within the PLEXE). In most cases, CAVs on the mixed lane are ahead of the target platoons when they are selected as joiners. That is why the joiner briefly decelerates to allow the target platoon to pass. Once the last platoon member has moved ahead, the joiner initiates a lane change and accelerates to catch up with the platoon. The joiner slows down again to synchronize with the platoon speed by adopting the CACC mechanism. 

\subsubsection{Joining quality}
\begin{figure}
\centering
\begin{subfigure}{0.8\linewidth}
    \centering
    \includegraphics[width=\linewidth]{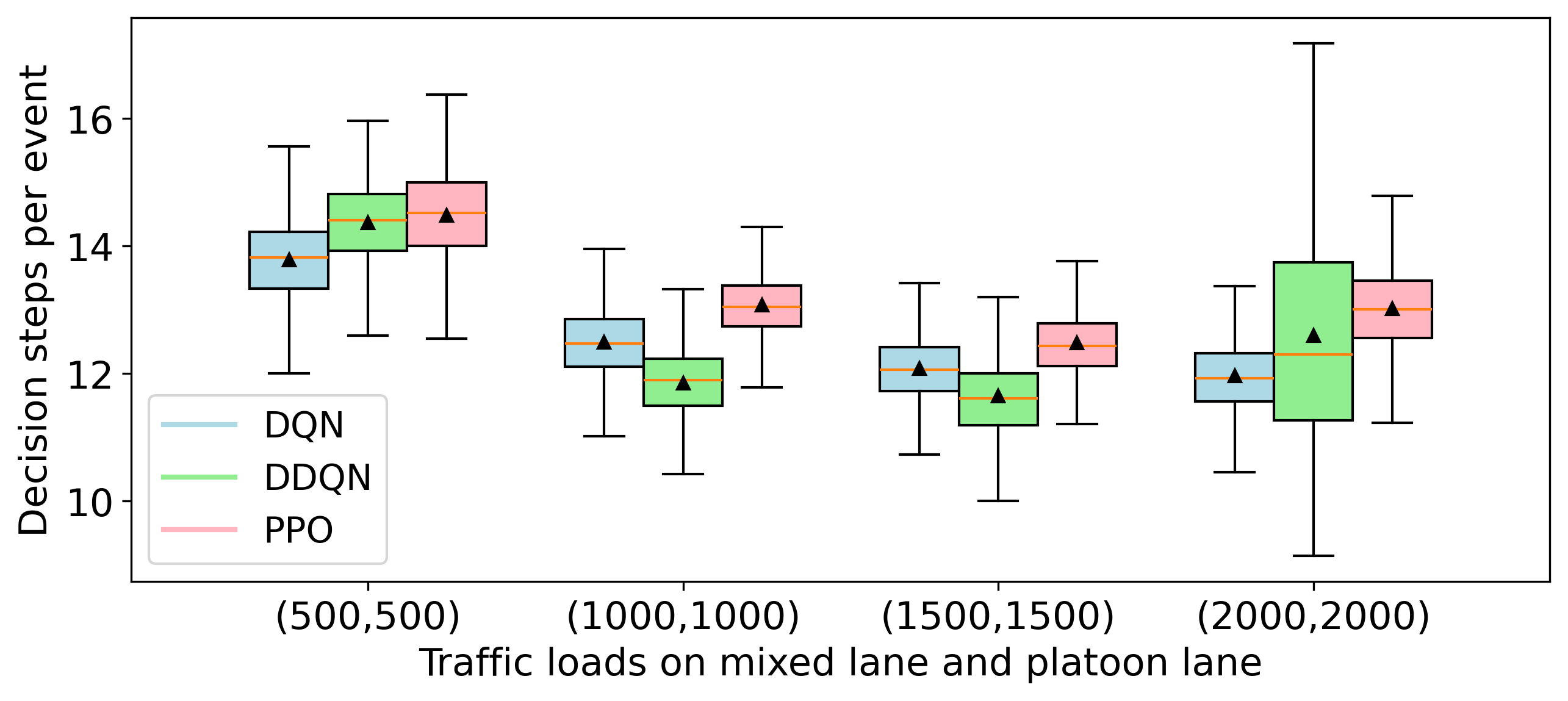}
    \caption{Symmetric loads}
\end{subfigure}
\begin{subfigure}{0.8\linewidth}
    \centering
    \includegraphics[width=\linewidth]{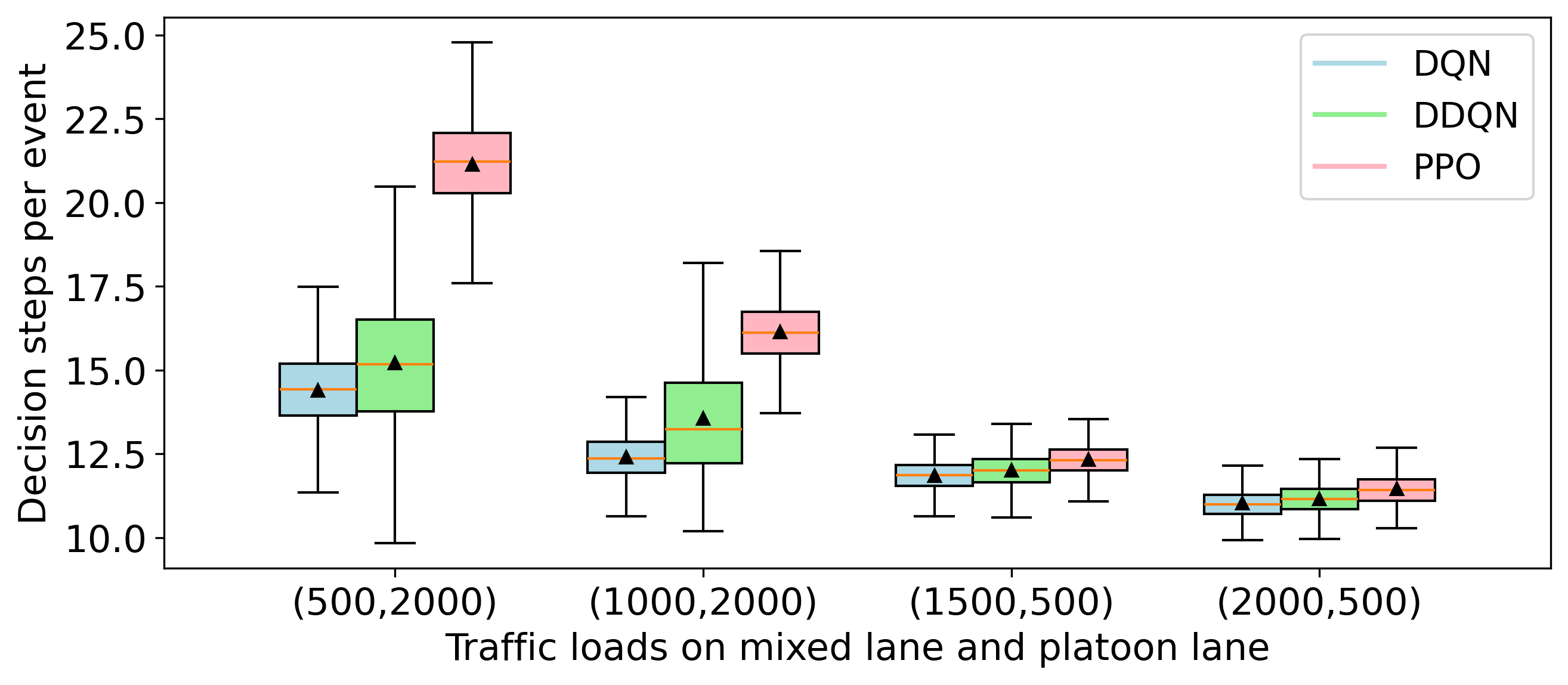}
    \caption{Asymmetric loads}
\end{subfigure}
\caption{Number of decision steps per event.}
\label{fig:decision_steps}
\end{figure}

We first investigate joining quality regarding the average number of decision steps until completing the joining, as shown in Fig. \ref{fig:decision_steps}. Boxplots use the standard 1.5×IQR whisker definition. Outlier markers are omitted for clarity. As previously shown in Fig. \ref{fig:learning_metrics}(d) PPO has more decision steps per event than the other two algorithms in convergence, we show in detail the distributions for different traffic load levels. With higher symmetric traffic loads in Fig. \ref{fig:decision_steps} (a), the number of decision steps gets smaller, especially for the DDQN. This may be explained by the decreased vehicle speeds due to high traffic density on both lanes. In cases with asymmetric loads (see Fig. \ref{fig:decision_steps} (b)), PPO takes many more decision steps for platoon joining than DQN and DDQN do when the traffic on the platoon lane is much more than that on the mixed lane, i.e., (500, 2000) and (1000, 2000). This difference mitigates a lot for the opposite demand cases. This implies that PPO's control of platoon joining is more conservative and dynamic when the platoon lane has much higher traffic density than the mixed lane. To some extent, DQN and DDQN both reflect their joining possibilities within stable decision steps when meeting largely asymmetric traffic loads on highways.

\begin{figure}
\centering
\begin{subfigure}{0.8\linewidth}
    \includegraphics[width=\linewidth]{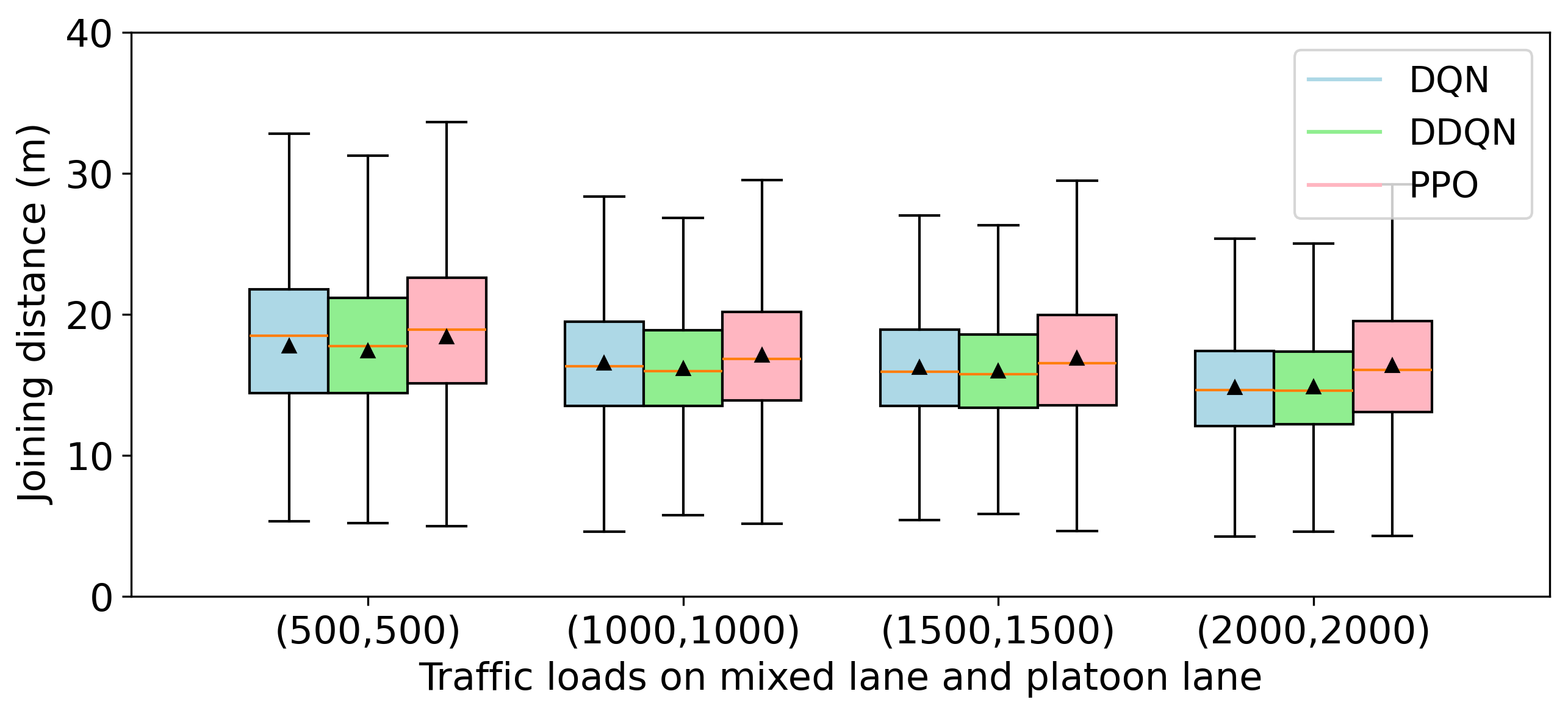}
    \caption{Symmetric loads}
\end{subfigure}
\begin{subfigure}{0.8\linewidth}
    \includegraphics[width=\linewidth]{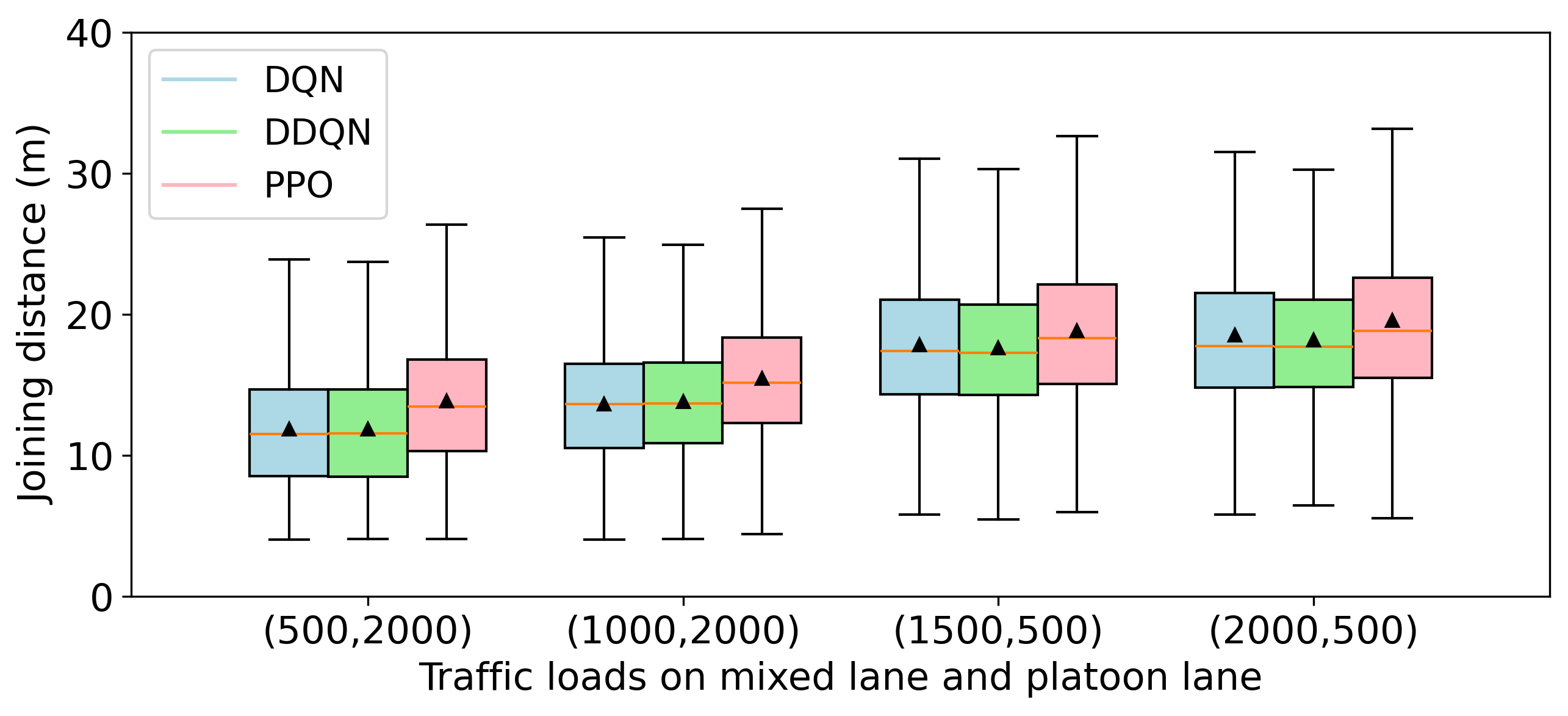}
    \caption{Asymmetric loads}
\end{subfigure}
\caption{Joining distance between $V_e$ and $V_{target}$}
\label{fig:join_dist}
\end{figure}

We then check the joining distance between the joiner and its vehicle ahead (i.e., the last member of the target platoon), as shown in Fig. \ref{fig:join_dist}. As a whole, the distributions of average joining distances seem very stable in symmetric loads, and they are located between 10 and 20 meters. In asymmetric loads, all three algorithms generate a bit longer distances for the cases of low traffic on the platoon lane, i.e., (1500, 500) and (2000, 500). No matter what, the observations are aligned with our previous stable-maneuver setting, where no penalty is valued in this range during the training procedure (see Eq. (\ref{eq: unstable_penalty})).  

\subsubsection{Robustness analysis}
\label{sec:robustness}
\begin{figure*}
\centering
\begin{subfigure}{\linewidth}
    \centering
    \includegraphics[width=\linewidth]{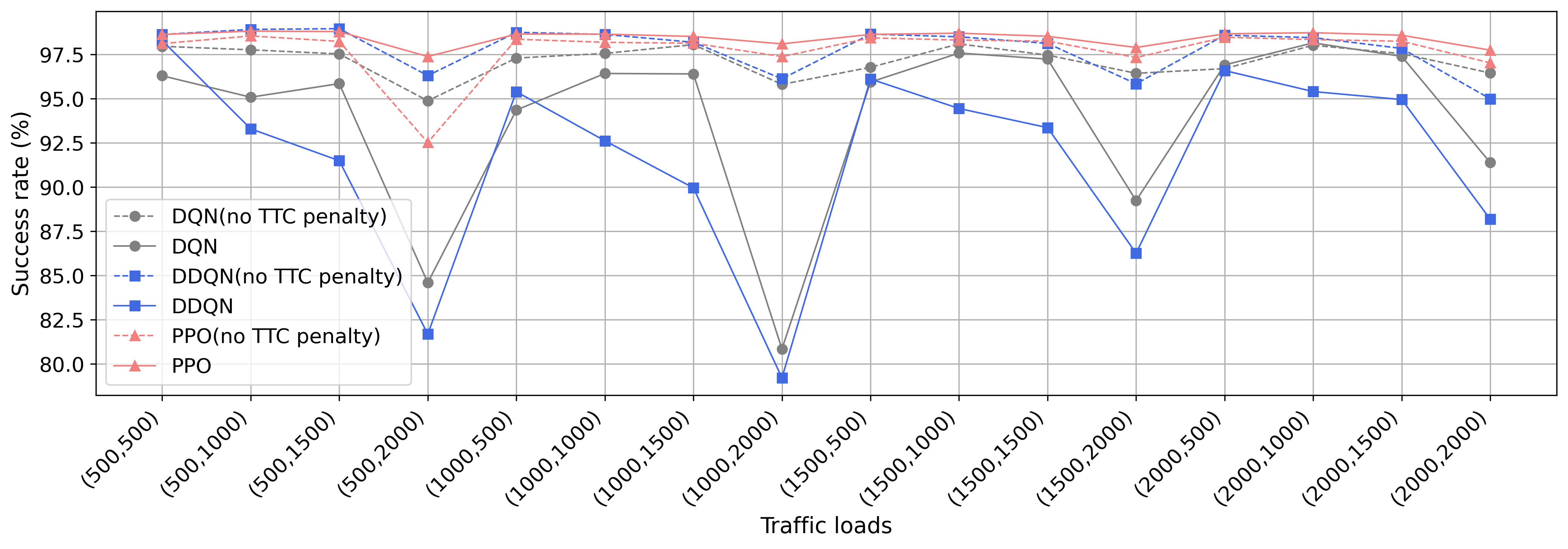}
    \caption{Success rate}
\end{subfigure}
\hfill
\begin{subfigure}{\linewidth}
    \centering
    \includegraphics[width=\linewidth]{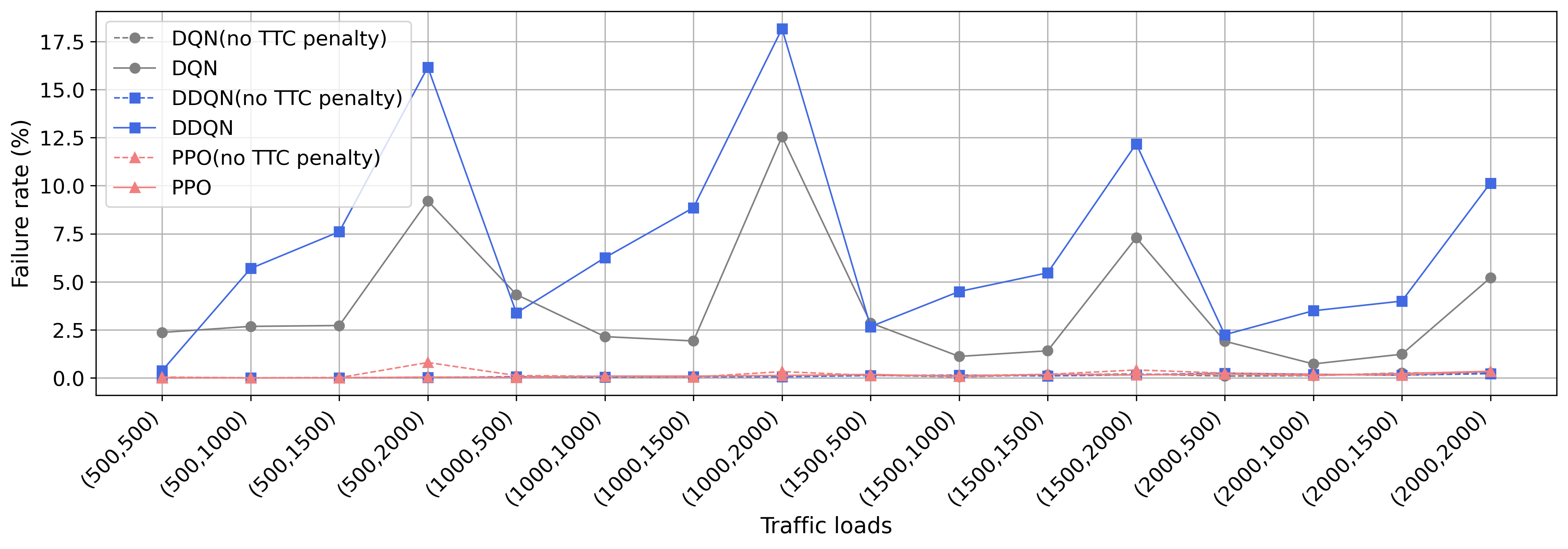}
    \caption{Failure rate}
\end{subfigure}
\hfill
\begin{subfigure}{\linewidth}
    \centering
    \includegraphics[width=\linewidth]{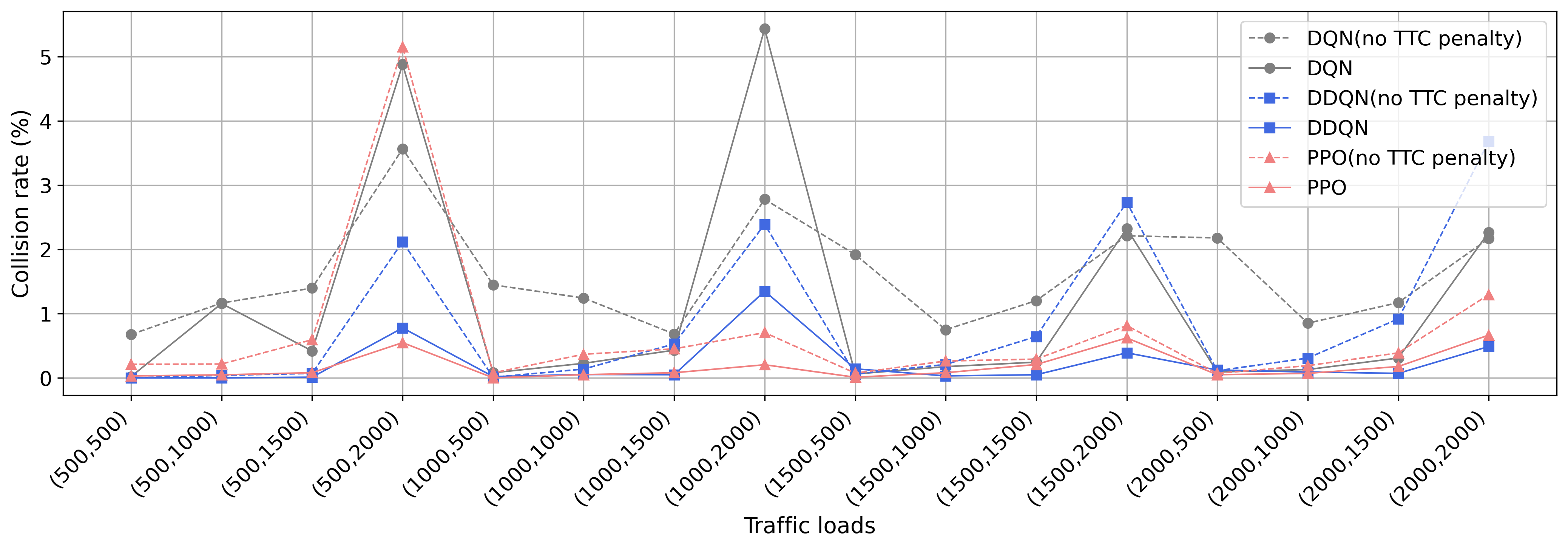}
    \caption{Collision rate}
\end{subfigure}
\caption {Model tests of algorithms with or without TTC penalty ($r_r$) in different traffic loads.}
\label{fig:robustness}
\end{figure*}

The success, failure, and collision percentages of platoon joining under different traffic loads were recorded to assess the robustness of the three types of deep RL approaches, as shown in Fig. \ref{fig:robustness}. The solid lines refer to the performance of the designed algorithms, whereas the dashed lines are those without taking the TTC penalty (i.e., $r_r$ in Eq. (\ref{eq: risky_penalty})) into account. Obviously, PPO presents the best performance with more than 97.5\% success rate (see Fig. \ref{fig:robustness}(a)) and less than 1.0\% collision rate (see Fig. \ref{fig:robustness}(c)) for all levels of traffic loads. Unexpectedly‌, the success rates of DDQN and DQN drop a lot, even under 90.0\% in the four traffic demand pairs of (500, 2000), (1000, 2000), (1500, 2000), and (2000, 2000). This can be explained by the large rates of failed joinings (including wrong joining positions and uncompleted joinings) due to the platoon lane with the highest traffic of 2000 veh/h (see Fig. \ref{fig:robustness}(b)). Compared to the variants of the three algorithms without TTC penalty, there are no surprisingly large failures because of the removed TTC constraints. However, this causes most of their collision rates to be relatively larger than those of the designed algorithms with the TTC risky penalty, as shown in Fig. \ref{fig:robustness}(c). Based on these results, the policy-based PPO algorithm seems more robust than the value-based DDQN and DQN algorithms, and PPO is even better when the TTC risky penalty is included in the reward function. 

\subsection{Safety shield for lane changes}
\label{subsec_safety_shield}
The importance of the risky maneuver penalty designed in the reward function was demonstrated above. Differently regarding this "inner" safety mode, we also intend to investigate external protection when the algorithms are designed without the inner TTC conditions, causing relatively high collision rates for the high traffic demands on the platoon lane (see Fig. \ref{fig:robustness}(c)). A safety shield (noted as "SS") beyond the lane-changing decisions made by the deep RL algorithms was implemented. The rear and front time-to-collision (TTC) conditions are both applied before platoon joining or speed adjustment on the mixed lane, following a decision by the learning algorithm, with the same settings as in the reward function design. Specifically, they are $\text{min}(TTC_{(V_e, V_{lf})},\allowbreak TTC_{(V_e, V_{target})}) >=2$ for the lane-change execution; and 2) $\text{min}(TTC_{(V_e, V_{fm})},\allowbreak TTC_{(V_e, V_{pm})})\allowbreak >=1$ for speed adjustement actions. The obtained results of metrics are given in Table \ref{tab:safety_shield}. It shows that collision rates are almost reduced to zero ($<$ 0.001), and they are transferred as abandonment. Notice that very small parts of previous successful joinings are also abandoned. The abandoned joinings are related to previous dangerous but successful joinings with a small TTC (i.e., $<$ 2 s). Again, PPO shows relatively higher success joining rates ($>=$ 96.8\%) with higher density traffic, followed by DDQN, which even demonstrates the best joining efficiency reaching 97.1\% in the asymmetric traffic case of (500, 2000). Mention that the safe PPO approaches proposed in \cite{krasowski2020} only achieved lane-changing goals of 75\% to 88\% with free collisions.

\begin{table*}
\centering
\caption{Joining performance of algorithms after applying a safety shield (SS)}
\label{tab:safety_shield}
\footnotesize
\begin{tabular}{cccccc}
\hline
& Metrics (\%) 
& \shortstack{(500, 2000)} 
& \shortstack{(1000, 2000)} 
& \shortstack{(1500, 2000)} 
& \shortstack{(2000, 2000)} \\
\hline
\multirow{4}{*}{\shortstack{DQN\\(SS instead of TTC penalty)}}&Success      & 93.7 & 93.3 & 92.8 & 91.6 \\
&Failure      & 0.2 & 0.2 & 0.2 & 0.1 \\
&Collision    & 0.0 & 0.0 & 0.0 & 0.0 \\
&Abandonment  & 6.1 & 6.5 & 7.0 & 8.3 \\
\hline
\multirow{4}{*}{\shortstack{DDQN\\(SS instead of TTC penalty)}}&Success      & 97.1 & 96.3 & 96.3 & 95.5 \\
&Failure      & 0.1 & 0.3 & 0.1 & 0.2 \\
&Collision    & 0.0 & 0.0 & 0.0 & 0.0 \\
&Abandonment  & 2.8 & 3.4 & 3.6 & 4.3 \\
\hline
\multirow{4}{*}{\shortstack{PPO\\(SS instead of TTC penalty)}}&Success   & 94.4  & 96.1  & 96.9  & 96.8  \\
&Failure      &  0.4 & 0.4  & 0.3 & 0.3  \\
&Collision    &  0.0 & 0.0 & 0.0 & 0.1  \\
&Abandonment  & 5.2 & 3.5 & 2.8  & 2.8  \\
\hline
\end{tabular}
\end{table*} 

\section{Conclusion}
In this paper, we proposed a deep RL-based framework for optimizing platoon joining on highways with mixed traffic. We primarily implemented the typical value-based DQN and DDQN algorithms and the policy-based PPO algorithm for this control. Compared to the DQN and DDQN, the PPO achieves superior performance on model training and testing, although it takes a few more decision steps (about +10\%) for platoon joining and longer distances (+~0.9~$\sim$~1.2~m.) to target platoons on average for the model tests. Concretely, the PPO performs excellently in all cases of traffic load testing, with about 98\% of successful joinings and less than 1\% for both failures and collisions. Regarding some quite degraded results of DQN and DDQN when the platoon load is set to 2,000 veh/h where a very short traffic head time is generated, we adopted strict conditions set by the rear/front TTC constraints before implementing the joining decisions that were learned by our approaches. The success rates under this safety shield for those specific cases are eventually over 92\% for DQN and 96\% for DDQN with guarantees of free collisions. 

Despite the generality of the proposed deep RL modeling framework for CAV platoon joining and results comparison by the proposed deep RL algorithms, a limitation of this study is related to the definite discrete action space for speed adjustments and its lack of optimization. Although driving comfort was considered through the inclusion of a jerk penalty in the reward function, joining efficiency is not confirmed as the ego vehicle's speed trajectory is not optimized explicitly. Future work will focus on enhancing the reward design by incorporating speed trajectory optimization objectives, such as energy efficiency and time-efficient platoon joining globally. Furthermore, the proposed framework will be extended by developing an efficient strategy for CAVs to safely and effectively exit platoons.

\newpage
\appendix
\counterwithin{table}{section}
\renewcommand{\thetable}{\Alph{section}.\arabic{table}}

\section{Configuration}

\begin{table}[h]
    \caption{Configuration for the DQN and DDQN algorithms}
    \label{tab:parameter_DQN_DDQN}
    \scriptsize
\begin{tabularx}{\linewidth}{l X}
        \hline\noalign{\smallskip}
        \textbf{Hyperparameter} & \textbf{Value/Definition} \\
        \hline\noalign{\smallskip}
        Model type & Fully connected online/target neural networks (MLP) \\ 
        Hidden layers & 4 \\ 
        Hidden units & $[$32, 64, 64, 32$]$ \\ 
        Activation function & LeakyReLU \\ 
        Output layer & Linear \\ 
        Input dimension & 15 \\ 
        Output dimension & 6 \\ 
        Optimizer & Adam \\ 
        Loss function & SmoothL1Loss \\ 
        Learning rate $\alpha$  & 0.0002 \\  
        Discount factor $\gamma$ & 0.9 \\  
        Soft update coefficient $\tau$ & 0.001 \\
        Soft update frequency $C$ & $10^3$ \\
        Decay($\epsilon$) & Exponential decay \\  
        $\epsilon_\text{start}$ & 1 \\  
        $\epsilon_\text{min}$ & 0.01 \\  
        $\epsilon_\text{decay}$ & $6\times10^5$ \\  
        Mini-batch size $M$ & 64 \\  
        Replay buffer size & $2\times10^4$ \\   
        Training warm-up steps & $1.2\times10^3$ \\    
        Maximum training steps & $10^6$ \\ 
        Maximum steps per episode & $10^3$ \\
        Decision step duration & 0.5s (5 simulation steps) \\
        \hline\noalign{\smallskip}
\end{tabularx}
\end{table}
\begin{table}[h]
    \caption{Configuration for the PPO algorithm}
    \label{tab:parameter_PPO}
    \scriptsize
\begin{tabularx}{\linewidth}{l X}
        \hline\noalign{\smallskip}
        \textbf{Hyperparameter} & \textbf{Value/Definition} \\
        \hline\noalign{\smallskip}
        Model type & Fully connected actor/critic neural networks (MLP) \\ 
        Hidden layers & 4 \\ 
        Hidden units & $[$32, 64, 64, 32$]$ \\ 
        Activation function & LeakyReLU \\ 
        Output layer & Linear \\ 
        Input dimension (actor network) & 15 \\ 
        Output dimension (actor network) & 6 \\ 
        Input dimension (critic network) & 15 \\ 
        Output dimension (critic network) & 1 \\ 
        Optimizer & Adam \\ 
        Actor learning rate $\alpha_{\text{actor}}$  & 0.0001 \\ 
        critic learning rate $\alpha_{\text{critic}}$  & 0.001 \\  
        Discount factor $\gamma$ & 0.9 \\    
        Clipping coefficient $\epsilon_{\text{clip}}$ & 0.1 \\ 
        GAE coefficient $\lambda$ & 0.95\\
        Critic loss coefficient $c_1$ & 0.5\\
        Entropy coefficient $c_2$ & 0.01\\
        Epochs $K$ & 5\\
        Mini-batch size $M$ & 256 \\ 
        Rollout size & 2048 \\
        Training warm-up steps & $1.2\times10^3$ \\    
        Maximum training steps & $10^6$ \\ 
        Maximum steps per episode & $10^3$ \\
        Decision step duration & 0.5s (5 simulation steps) \\
        \hline\noalign{\smallskip}
\end{tabularx}
\end{table}

\begin{table}[h]
    \caption{Configuration for traffic scenarios}
    \label{tab:parameter_scenarios}
    \scriptsize
\begin{tabularx}{\linewidth}{l X}
        \hline\noalign{\smallskip}
        \textbf{Hyperparameter} & \textbf{Value/Definition} \\
        \hline\noalign{\smallskip}
        Traffic load per lane & Varying from 500 veh/h to 2,000 veh/h, with an increment of 500 veh/h \\  
        CAV penetration rate \\(on the mixed lane) & 50\% \\  
        Communication radius $R$ & 200 m \\
        Human driver imperfection & 0.5 \\  
        Rear distance search $\delta_1$ & 150 m \\ 
        Front distance search $\delta_2$ & 5 m \\ 
        Intervehicle distance (CACC) & 5 m \\  
        Speed limits on lane & $v_{m}$ = 100 km/h; $v_{p}$ = 120 km/h \\
        \hline\noalign{\smallskip}
\end{tabularx}
\end{table}

\end{document}